\documentclass[10pt,twocolumn,letterpaper]{article}

\usepackage{iccv}
\usepackage{times}
\usepackage{epsfig}
\usepackage{graphicx}
\usepackage{amsmath}
\usepackage{amssymb}
\usepackage{multirow}
\usepackage{threeparttable}
\usepackage{stmaryrd}
\usepackage{array}
\usepackage{booktabs}
\usepackage{mathabx}
\usepackage{authblk}
\usepackage[breaklinks=true,bookmarks=false]{hyperref}

\iccvfinalcopy % *** Uncomment this line for the final submission

\def\iccvPaperID{****} % *** Enter the ICCV Paper ID here

\ificcvfinal\fi

\begin{document}

%%%%%%%%% TITLE
\title{M$^{2}$FPA: A Multi-Yaw Multi-Pitch High-Quality Dataset and Benchmark for Facial Pose Analysis}

\author{Peipei Li$^{1,2}$, Xiang Wu$^{1}$,
Yibo Hu$^{1}$,
Ran He$^{1,2}$\thanks{corresponding author} ,
Zhenan Sun$^{1,2}$\\

$^{1}$CRIPAC$\;\&\;$NLPR\;\&\;CEBSIT, CASIA  \quad $^{2}$University of Chinese Academy of Sciences\\
%$^{1}$National Laboratory of Pattern Recognition, CASIA, Beijing, China 100190\\
%$^{2}$Center for Research on Intelligent Perception and Computing, CASIA, Beijing, China 100190\\
%$^{3}$University of Chinese Academy of Sciences, Beijing, China 100049\\

Email: \{peipei.li, yibo.hu\}@cripac.ia.ac.cn,  alfredxiangwu@gmail.com, \{rhe, znsun\}@nlpr.ia.ac.cn\\

\color{magenta}{\url{https://pp2li.github.io/M2FPA-dataset/}}
}

\maketitle
% Remove page # from the first page of camera-ready.
\ificcvfinal\thispagestyle{empty}\fi

%%%%%%%%% ABSTRACT
\begin{abstract}
  Facial images in surveillance or mobile scenarios often have large view-point variations in terms of pitch and yaw angles. These jointly occurred angle variations make face recognition challenging. Current public face databases mainly consider the case of yaw variations. In this paper, a new large-scale Multi-yaw Multi-pitch high-quality
database is proposed for Facial Pose Analysis (M$^2$FPA), including face frontalization, face rotation, facial pose estimation and pose-invariant face recognition. It contains 397,544 images of 229 subjects with yaw, pitch, attribute, illumination and accessory. M$^2$FPA is the most comprehensive multi-view face database for facial pose analysis. Further, we provide an effective benchmark for face frontalization and pose-invariant face recognition on M$^2$FPA with
several state-of-the-art methods, including DR-GAN\cite{tran2017disentangled}, TP-GAN\cite{huang2017beyond} and CAPG-GAN\cite{hu2018pose}. We believe that the new database and benchmark can significantly push forward the advance of facial pose analysis in real-world applications. Moreover, a simple yet effective parsing guided discriminator is introduced to capture the local consistency during GAN optimization. Extensive quantitative and qualitative results on M$^2$FPA and Multi-PIE demonstrate the superiority of our face frontalization method. Baseline results for
both face synthesis and face recognition from state-of-the-art methods demonstrate the challenge offered by this new
database.
\end{abstract}

%%%%%%%%% BODY TEXT
\section{Introduction}

With the development of deep learning, face recognition systems have achieved 99\% accuracy ~\cite{Parkhi2015DeepFR,Cao2018VGGFace2AD,wu2018light} on some popular databases~\cite{huang2008labeled,KemelmacherShlizerman2016TheMB}. However, in some realworld surveillance or mobile scenarios, the captured face images often contain extreme view-point variations so that face recognition performance is significantly affected. Recently, the great progress of face synthesis~\cite{hu2018pose,huang2017beyond,zhao2017dual}  has
pushed forward the development of ¡°recognition via generation¡±. TP-GAN~\cite{huang2017beyond} and CAPG-GAN~\cite{hu2018pose} perform face frontalization to improve recognition accuracy under large poses. DA-GAN~\cite{zhao2017dual} is proposed to simulate profile face
images, facilitating pose-invariant face recognition. However, their performance often depends on the diversity of pose variations in the training databases.

The existing face databases with pose variations can be categorized into two classes. The ones, such as LFW~\cite{huang2008labeled}, IJB-A~\cite{klare2015pushing} and VGGFace2~\cite{Cao2018VGGFace2AD}, are collected from the Internet, whose pose variations follow a long-tailed distribution. Moreover, it is obvious that obtaining the accurate pose labels is difficult for these databases. The others, including CMU PIE~\cite{sim2002cmu}, CAS-PEAL-R1~\cite{gao2008cas} and CMU Multi-PIE~\cite{gross2010multi}, are captured under the constrained environment across accurate poses. These databases often pay attention to yaw angles without considering pitch angles. However, facial images captured in surveillance or mobile scenarios often have large yaw and pitch variations simultaneously. Such the face recognition across both yaw and pitch angles needs to be extensively evaluated in order to ensure the robustness of recognition system. Therefore, it is
crucial to provide researchers with a multi-yaw multi-pitch high-quality face database for facial pose analysis, including face frontalization, face rotation, facial pose estimation and pose-invariant face recognition.

In this paper, a \textbf{M}ulti-yaw \textbf{M}ulti-pitch high-quality database for \textbf{F}acial \textbf{P}ose \textbf{A}nalysis (M$^2$FPA) is proposed to address this issue. The comparisons with the existing facial pose analysis databases are summarized in Table \ref{tab:database}. The main advantages lie in the following aspects: (1) \textbf{Large-scale.} M$^2$FPA includes totally 397,544 images of 229 subjects with 62 poses, 4 attributes and 7 illuminations. (2) \textbf{Accurate and diverse poses.} We design an acquisition system to simultaneously capture 62 poses, including 13 yaw angles (ranging from $-90^\circ$ to $+90^\circ$), 5 pitch angles (ranging from $-30^\circ$ to $+45^\circ$) and 44 yaw-pitch angles. (3) \textbf{High-resolutions.} All the images are captured by the SHL-200WS (2.0-megapixel CMOS camera), which leads to high-quality resolutions ($1920\times1080$). (4) \textbf{Accessory.} We use five types of glasses as accessories to further increase the diversity of our database with occlusions.

\begin{table*}[htbp]\scriptsize
\centering
\caption{Comparisons of existing facial pose analysis databases.  Image Size is the average size across all the images in the database. $^\star$In Multi-PIE, part of frontal images are 3072$\times$2048 in size, but the most are 640$\times$480 resolution. $^\dotplus$Images have much background in IJB-A.}
\resizebox{0.99\textwidth}{!} {
\begin{tabular}{lcccccccccccc}
\hline
\cline{1-13}
\textbf{Database}&\textbf{Yaw}&\textbf{Pitch} &\textbf{Yaw-Pitch} &\textbf{Attributes} &\textbf{Illuminations}  & \textbf{Subjects} &\textbf{Images}& \textbf{Image Size}&\textbf{Controllabled}&\textbf{Size[GB]}&\textbf{Paired}&\textbf{Year}\\
\hline
PIE~\cite{sim2002cmu}&9&2&2&4 &21&68&41,000+&640$\times$486&$\checkmark$&40&$\checkmark$&2003\\
LFW~\cite{huang2008labeled}&No label&No label&No label&No label&No label&5,749&13,233&250$\times$250&$\bigtimes$&0.17&$\bigtimes$&2007\\
CAS-PEAL-R1~\cite{gao2008cas}&7&2&12&5&15&1,040&30,863&640$\times$480 &$\checkmark$&26.6&$\checkmark$&2008\\
Multi-PIE \cite{gross2010multi}&13&0&2&6&19&337&755,370&640$\times$480$^ \star$&$\checkmark$&305&$\checkmark$&2009\\
IJB-A~\cite{klare2015pushing}&No label&No label&No label&No label&No label&500&25,809&1026$\times$698$^\dotplus$&$\bigtimes$&14.5&$\bigtimes$&2015\\
CelebA~\cite{liu2015deep}&No label&No label&No label&No label&No label&10,177&202,599&505$\times$606&$\bigtimes$&9.49&$\bigtimes$&2016\\
CelebA-HQ~\cite{karras2017progressive}&No label&No label&No label&No label&No label&No label&30,000&1024$\times$1024&$\bigtimes$&27.5&$\bigtimes$&2017\\
FF-HQ~\cite{karras2018style}&No label&No label& No label&No label&No label&No label&70,000&1024$\times$1024&$\bigtimes$&89.3&$\bigtimes$&2018\\
\hline
\textbf{M$^{2}$FPA} (Ours)&\textbf{13}&\textbf{5}&\textbf{44}&\textbf{4}&\textbf{7}&\textbf{229}&\textbf{397,544}&\textbf{1920$\times$1080}&$\checkmark$&\textbf{421}&$\checkmark$&\textbf{2019}\\
\hline
\cline{1-13}
\end{tabular}
\label{tab:database}}
\vspace{-0.3cm}
\end{table*}

To the best of our knowledge, M$^2$FPA is the most comprehensive multi-view face database which covers variations in yaw, pitch, attribute, illumination and accessory. M$^2$FPA will provide researchers developing and evaluating the new algorithms for facial pose analysis, including face frontalization, face rotation, facial pose estimation and pose-invariant face recognition.
Furthermore, in order to provide an effective benchmark for face frontalization and pose-invariant face recognition on M$^2$FPA, we implement and evaluate several state-of-the-art methods, including DR-GAN\cite{tran2017disentangled}, TP-GAN\cite{huang2017beyond} and CAPG-GAN\cite{hu2018pose}.

In addition, we propose a simple yet effective parsing guided discriminator, which introduces parsing maps~\cite{liu2015deep} as a flexible attention to capture the local consistency during GAN optimization. First, a pre-trained facial parser captures the three local masks, including hairstyle, skin and facial features (eyes, nose and mouth). Second, we treat these parsing masks as the soft attention, facilitating the synthesized frontal images and the ground truth. Then, these local features are fed into a discriminator, called parsing guided discriminator, to ensure the local consistency of the synthesized frontal images. In this way, we can synthesize photo-realistic frontal images with extreme yaw and pitch variations on M$^2$FPA and Multi-PIE databases.

The main contributions of this paper are as follows:
\begin{itemize}
  \item We introduce a Multi-yaw Multi-pitch high-quality database for Facial Pose Analysis (M$^2$FPA). It contains 397,544 images of 229 subjects with yaw, pitch, attribute, illumination and accessory.
  \item We provide a comprehensive qualitative and quantitative benchmark of several state-of-the-art methods for face frontalization and pose-invariant face recognition, including DR-GAN\cite{tran2017disentangled}, TP-GAN\cite{huang2017beyond} and CAPG-GAN\cite{hu2018pose}, on M$^2$FPA.
  \item We propose a simple yet effective parsing guided discriminator, which introduces parsing maps as a soft attention to capture the local consistency during GAN optimization. In this way, we can synthesize photo-realistic frontal images on M$^2$FPA and Multi-PIE.
\end{itemize}

\begin{figure*}[t]
\setlength{\belowcaptionskip}{-0.2cm}
\begin{center}
\includegraphics[width=0.85\linewidth]{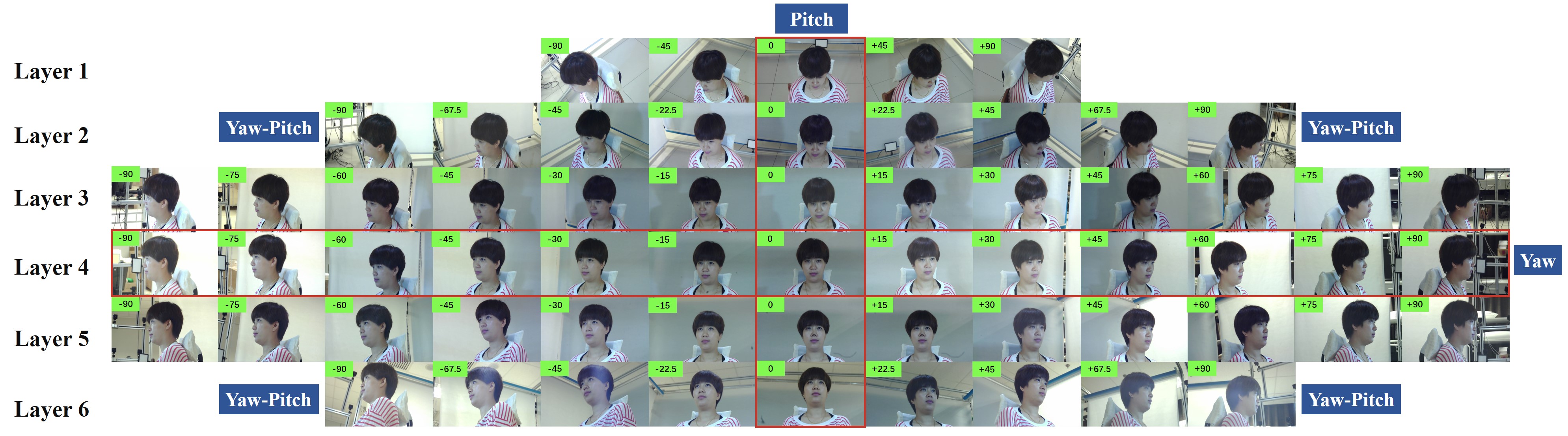}%images//GLA-GAN
\end{center}
\vspace{-0.4cm}
   \caption{An example of the yaw and pitch variations in our M$^{2}$FPA database. From top to bottom, the pitch angles of the 6 camera layers are $+45^\circ$, $+30^\circ$, $+15^\circ$, $0^\circ$, $-15^\circ$ and $-30^\circ$, respectively. The yaw pose of each image is shown in the green box.}\label{Fig:example}
\vspace{-0.4cm}
\end{figure*}

\section{Related Work}

\subsection{Databases}
The existing face databases with pose variations can be categorized into two classes. The ones, including LFW \cite{huang2008labeled}, IJB-A \cite{klare2015pushing}, VGGFace2~\cite{Cao2018VGGFace2AD}, CelebA \cite{liu2015deep} and CelebA-HQ \cite{karras2017progressive}, are often collected from the Internet. Therefore, the pose variations in these databases follow a long-tailed distribution, that is there are lots of nearly frontal faces but few profile ones. In addition, it is expensive to obtain the precious pose labels for these facial images, which leads to difficulties for face frontalization, face rotation and facial pose estimation.
The others, such as CMU PIE~\cite{sim2002cmu}, CMU Multi-PIE~\cite{gross2010multi} and CAS-PEAL-R1~\cite{gao2008cas}, are captured under constrained environment with precise controlling of angles. CUM PIE and CMU Multi-PIE have only yaw angles ranging from $-90^\circ$ to $90^\circ$. CAS-PEAL-R1 contains 14 yaw-pitch angles, but these pitch variations are captured by asking the subjects to look upward/downward, which leads to the inaccurate pose labels. Moreover, in CAS-PEAL-R1, only frontal facial images contain accessory variations. Different from these existing databases, M$^2$FPA contains variations including attribute, illumination, accessory across precious yaw and pitch angles.

\subsection{Face Rotation}
Face rotation is an extremely challenging ill-posed task in computer vision. In recent years, benefiting from Generative Adversarial Network (GAN)~\cite{goodfellow2014generative}, face rotation has made great progress. Currently, state-of-the-art face rotation algorithms can be categorized into two aspects, including 2D~\cite{tran2017disentangled,huang2017beyond,hu2018pose,zhao2018towards,shen2018faceid,tian2018cr} and 3D~\cite{yin2017towards,zhao2017dual,deng2018uv,zhao20183d,cao2018learning,moniz2018unsupervised} based methods.
For 2D based methods, Tran et.al \cite{tran2017disentangled} propose DR-GAN to disentangle pose variations from the facial images. TP-GAN \cite{huang2017beyond} employs a two path model, including global and local generators, to synthesize photo-realistic frontal faces.
Hu et. al \cite{hu2018pose} incorporate landmark heatmaps as a geometry guidance to synthesize face images with arbitrary poses.
PIM~\cite{zhao2018towards} performs face frontalization in a mutual boosting way with a dual-path generator. FaceID-GAN \cite{shen2018faceid} extends the conventional two-player GAN to three players, competing with the generator by disentangling the identities of real and synthesized faces.
Considering 3D-based methods, FF-GAN \cite{yin2017towards} incorporates 3DMM into GAN to provide the shape and appearance prior. DA-GAN~\cite{zhao2017dual} employs a dual architecture to refine a 3D simulated profile face. UV-GAN~\cite{deng2018uv} considers face rotation as a UV map completion task. 3D-PIM \cite{zhao20183d} incorporates a simulator with a 3D Morphable Model to obtain shape and appearance priors for face frontalization. Moreover, DepthNet~\cite{moniz2018unsupervised} infers plausible 3D transformations from one face pose to another, to realize face frontalization.

%-----------------------------------------
\section{The M$^{2}$FPA Database}
In this section, we present an overview of the M$^{2}$FPA database,  including how it was collected, cleaned, annotated and its statistics. To the best of our knowledge, M$^{2}$FPA is the first publicly available database that contains precise
and multiple yaw and pitch variations. In the rest of this section, we first introduce the hardware configuration and the data collection.  Then we describe the cleaning and annotating procedure. Finally, we present the statistics of M$^{2}$FPA, including the yaw and pitch variations, the types of attributes and the positions of illuminations.

\subsection{Data Acquisition}
\begin{figure}[t]
\setlength{\abovecaptionskip}{0cm}
\begin{center}
\includegraphics[width=0.8\linewidth]{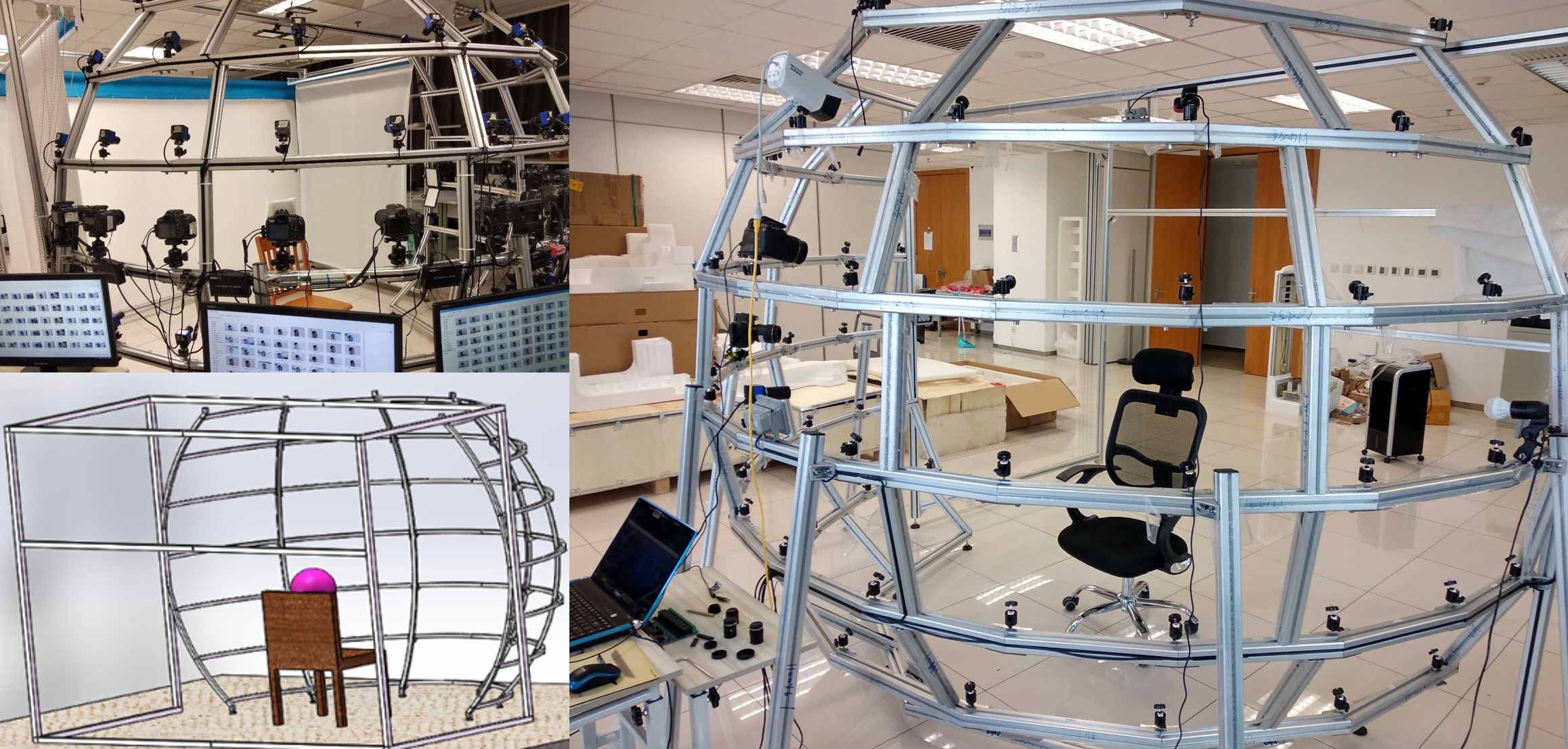}%images//GLA-GAN
\end{center}
\vspace{-0.4cm}
   \caption{Overview of the acquisition system. It contains total 7 horizontal layers. The bottom is the balanced layer and the rest are the camera layers.}\label{Fig:system_architecture}
\vspace{-0.4cm}
%\label{fig:synthesis_results}
\end{figure}

\begin{figure}[t]
\setlength{\abovecaptionskip}{0cm}
\begin{center}
\includegraphics[width=0.85\linewidth]{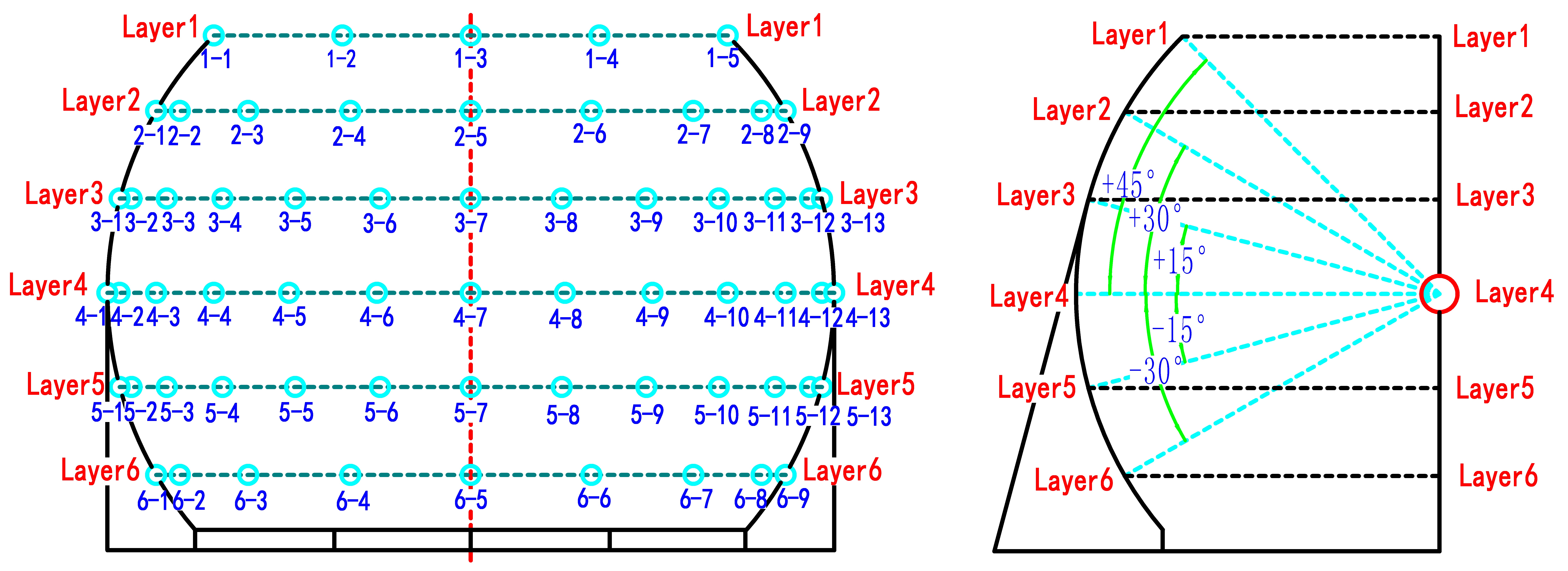}%images//GLA-GAN
\end{center}

\vspace{-0.4cm}
   \caption{The diagram of camera positions. The left and right are
the cutaways of frontal and side views, respectively. }\label{Fig:light}
\vspace{-0.4cm}

\end{figure}

We design a flexible multi-camera acquisition system to capture faces with multiple yaw and pitch angles. Figure \ref{Fig:system_architecture} shows an overview of the acquisition system. It is built by many removable brackets, forming an approximate hemisphere with a diameter of 3 meters. As shown in Figure \ref{Fig:light}, the acquisition system contains 7 horizontal layers, where the first six (Layer1$\sim$Layer6) are the camera layers and the last one is the balance layer. The interval between two adjacent layers is $15^\circ$. The Layer4 has the same height with the center of hemisphere (red circle in Figure \ref{Fig:light}). Therefore, we set the pitch angle of Layer4 to $0^\circ$. As a result, from top to bottom, the intervals between the rest 5 camera layers and the Layer4 are $+45^\circ$, $+30^\circ$, $+15^\circ$, $-15^\circ$ and $-30^\circ$, respectively.

\begin{figure}[t]
\setlength{\abovecaptionskip}{0cm}
\begin{center}
\includegraphics[width=0.85\linewidth]{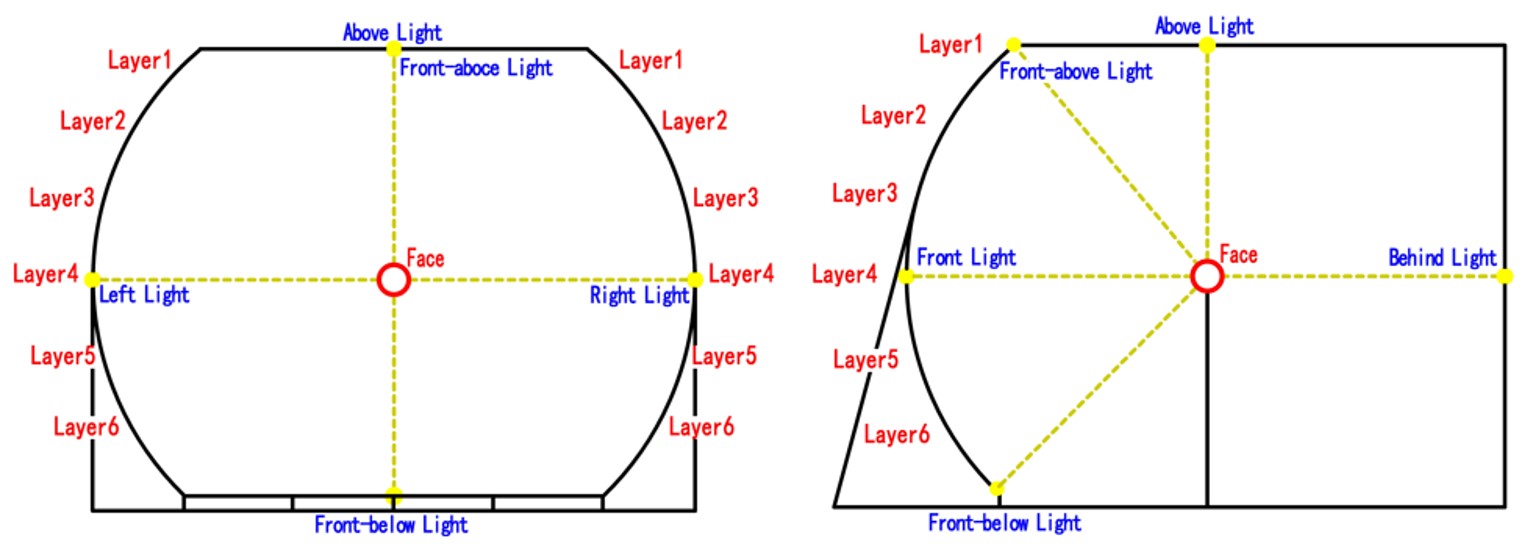}%images//GLA-GAN
\end{center}

\vspace{-0.4cm}
   \caption{The diagram of illumination positions. The left and right
are the cutaways of frontal and side views, respectively.}\label{Fig:position}
\vspace{-0.4cm}
\end{figure}

A total of 62 SHL-200WSs (2.0-megapixel CMOS camera with 12mm prime lens) are located on these 6 camera layers. As shown in Figure \ref{Fig:light}, there are 5, 9, 13, 13, 13 and 9 cameras on the Layer1, 2, 3, 4, 5 and 6, respectively. For each layer, the cameras are evenly located from $-90^\circ$ to $+90^\circ$. The detailed yaw and pitch angles of each camera can be found in Figure \ref{Fig:example} and Table \ref{tab:database111}. All
the 62 cameras are connected to 6 computers through USB
interfaces and a master computer synchronously dominates
these computers. We develop a software to simultaneously
control the 62 cameras and collect all the 62 images in one shot to ensure the consistency. In addition, as described in Figure \ref{Fig:position}, there are 7 different directions of light source equipped on our acquisition system, including above, front, front-above, front-below, behind, left and right. In order to maintain the consistency of the background, we construct some brackets and white canvas behind the acquisition system, as shown in the upper left corner in Figure \ref{Fig:system_architecture}.

\begin{figure}[t]
\setlength{\abovecaptionskip}{0cm}
\begin{center}
\includegraphics[width=0.85\linewidth]{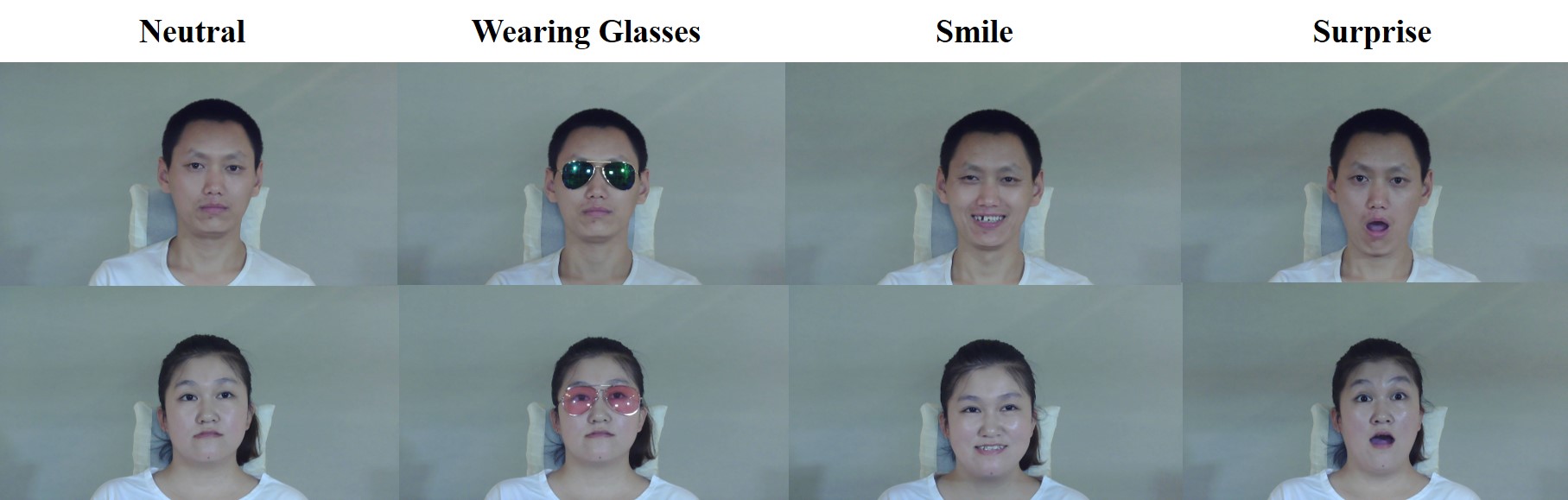}%images//GLA-GAN
\end{center}

\vspace{-0.4cm}
   \caption{Examples of four attributes in M$^{2}$FPA.}\label{Fig:example_expression}
\vspace{-0.4cm}
\end{figure}

A total of 300 volunteers are chosen to create the M$^{2}$FPA and all the participants have signed a license. During the collection procedure, we fix a chair and provide a headrest to ensure position of face is at the center of hemisphere. Each participant has 4 attributes, including neutral, wearing glass, smile and surprise. Figure \ref{Fig:example_expression} shows some examples of the attributes. Therefore, we totally capture $300 \times 62 \times 7 \times 4 = 520,800$ ($participants \times poses \times illuminations \times attributes$) facial images.

\subsection{Data Cleaning and Annotating}
After collection, we manually check all the facial images and remove those participants whose entire head is not captured by one or more cameras. In the end, we eliminate 71 participants with information missing, and the remaining 229 participants form our final M$^{2}$FPA database. Facial landmark detection is an essential preprocessing in facial pose analysis, such as face rotation and pose-invariant face recognition. However, current methods \cite{bulat2017far, DBLP:conf/cvpr/SunWT13} often fail to accurately detect facials landmarks with extreme yaw and pitch angles. In order to ease the utilization of our database, we manually mark the five facial landmarks of each image in M$^{2}$FPA.

\subsection{The Statistics of M$^{2}$FPA}
\renewcommand\arraystretch{1.3}
\begin{table}[htbp]\scriptsize
\centering
\caption{The poses, attributes and illuminations in M$^{2}$FPA.}
\begin{tabular}{ll|clll}
\toprule[1.1pt]
\multirow{11}{*}{\textbf{Poses}} &  \multicolumn{1}{|c|}{Pitch = $+45^\circ$ } &Yaw = $-90^\circ, -45^\circ, 0^\circ, +45^\circ, +90^\circ$\\
\cline{2-3}
&\multicolumn{1}{|l|}{\multirow{2}{*}{Pitch = $+30^\circ$  }}&Yaw = $-90^\circ, -67.5^\circ, -45^\circ, -22.5^\circ$\\
&\multicolumn{1}{|l|}{}& $0^\circ, +22.5^\circ, +45^\circ, +67.5^\circ, +90^\circ$ \\
\cline{2-3}
&\multicolumn{1}{|l|}{\multirow{2}{*}{Pitch = $+15^\circ$  }}&Yaw = $-90^\circ, -75^\circ, -60^\circ, -45^\circ, -30^\circ, -15^\circ$\\
&\multicolumn{1}{|l|}{}& $0^\circ, +15^\circ, +30^\circ, +45^\circ, +60^\circ, +75^\circ, +90^\circ$ \\
\cline{2-3}
&\multicolumn{1}{|l|}{\multirow{2}{*}{Pitch = $0^\circ$  }}&Yaw = $-90^\circ, -75^\circ, -60^\circ, -45^\circ, -30^\circ, -15^\circ$\\
&\multicolumn{1}{|l|}{}& $0^\circ, +15^\circ, +30^\circ, +45^\circ, +60^\circ, +75^\circ, +90^\circ$ \\
\cline{2-3}
&\multicolumn{1}{|l|}{\multirow{2}{*}{Pitch = $-15^\circ$  }}&Yaw = $-90^\circ, -75^\circ, -60^\circ, -45^\circ, -30^\circ, -15^\circ$\\
&\multicolumn{1}{|l|}{}& $0^\circ, +15^\circ, +30^\circ, +45^\circ, +60^\circ, +75^\circ, +90^\circ$ \\
\cline{2-3}
&\multicolumn{1}{|l|}{\multirow{2}{*}{Pitch = $-30^\circ$  }}&Yaw = $-90^\circ, -67.5^\circ, -45^\circ, -22.5^\circ$\\
&\multicolumn{1}{|l|}{}& $0^\circ, +22.5^\circ, +45^\circ, +67.5^\circ, +90^\circ$ \\
\midrule[0.8pt]
\multicolumn{2}{c|}{\textbf{Attributes}}& Happy, Normal, Wear glasses, Surprise\\
\midrule[0.8pt]
\multicolumn{2}{c|}{\multirow{2}{*}{\textbf{Illuminations}}}& Above, Front, Front-above, Behind\\
&&Front-below, Left, Right\\
\bottomrule[1.1pt]
\end{tabular}\label{tab:database111}
\vspace{-0.4cm}
\end{table}

After manually cleaning, we retain 397,544 facial images of 229 subjects, covering 62 poses, 4 attributes and 7 illuminations. Table \ref{tab:database111} presents the poses, attributes and illuminations of our M$^{2}$FPA database. Compared with the existing facial pose analysis databases, as summarized in Table \ref{tab:database}, the main advantages of M$^{2}$FPA lie in four-folds:

\begin{itemize}
    \item \textbf{Large-scale.} M$^{2}$FPA contains total 397,544 facial images of 229 subjects with 62 poses, 4 attributes and 7 illuminations. It spends almost one year to establish the multi-camera acquisition system and collect such a number of images.

    \item \textbf{Accurate and diverse poses.} Our acquisition system
can simultaneously capture 62 poses in one shot, including 13 yaw angles (ranging from $-90^\circ$ to  $+90^\circ$), 5 pitch angles (ranging from  $-30^\circ$ to  $+45^\circ$) and 44 yaw-pitch angles. To the best of our knowledge, M$^{2}$FPA is the first publicly available database that contains precise and multiple yaw and pitch angles.

    \item \textbf{High-resolution.} All the images are captured by the SHL-200WS (2.0-megapixel CMOS camera), leading to high resolution ($1920 \times 1080$).

    \item \textbf{Accessory.} In order to further increase the diversity of M$^{2}$FPA, we add five types of glasses as
the accessories, including dark sunglasses, pink sunglasses, round glasses, librarian glasses and rimless glasses.
\end{itemize}

%-----------------------------------------

\section{Approach}

In this section, we propose a parsing guided local discriminator into GAN training, as is shown in Figure~\ref{Fig:architecture_method}. We introduce parsing maps~\cite{liu2015deep} as a flexible attention to capture the local consistency of the real and synthesized frontal images. In this way, our method can effectively frontalize a face with yaw-pitch variations and accessory occlusions on the new M$^2$FPA database.

\subsection{Network Architecture}

Given a profile facial image $X$ and its corresponding frontal face $Y$, we can obtain the synthesized frontal image $\hat{Y}$ by a generator ${G_{{\theta _G}}}$,
\begin{equation}
\begin{array}{c}
\hat{Y} = {G_{{\theta _G}}}\left( X \right)
\end{array}
\end{equation}
where ${\theta _G}$ is the parameter of ${G_{{\theta _G}}}$. The architecture of generator is detailed in Supplementary Materials.

\begin{figure}[t]
\setlength{\abovecaptionskip}{0cm}
\begin{center}
\includegraphics[width=0.95\linewidth]{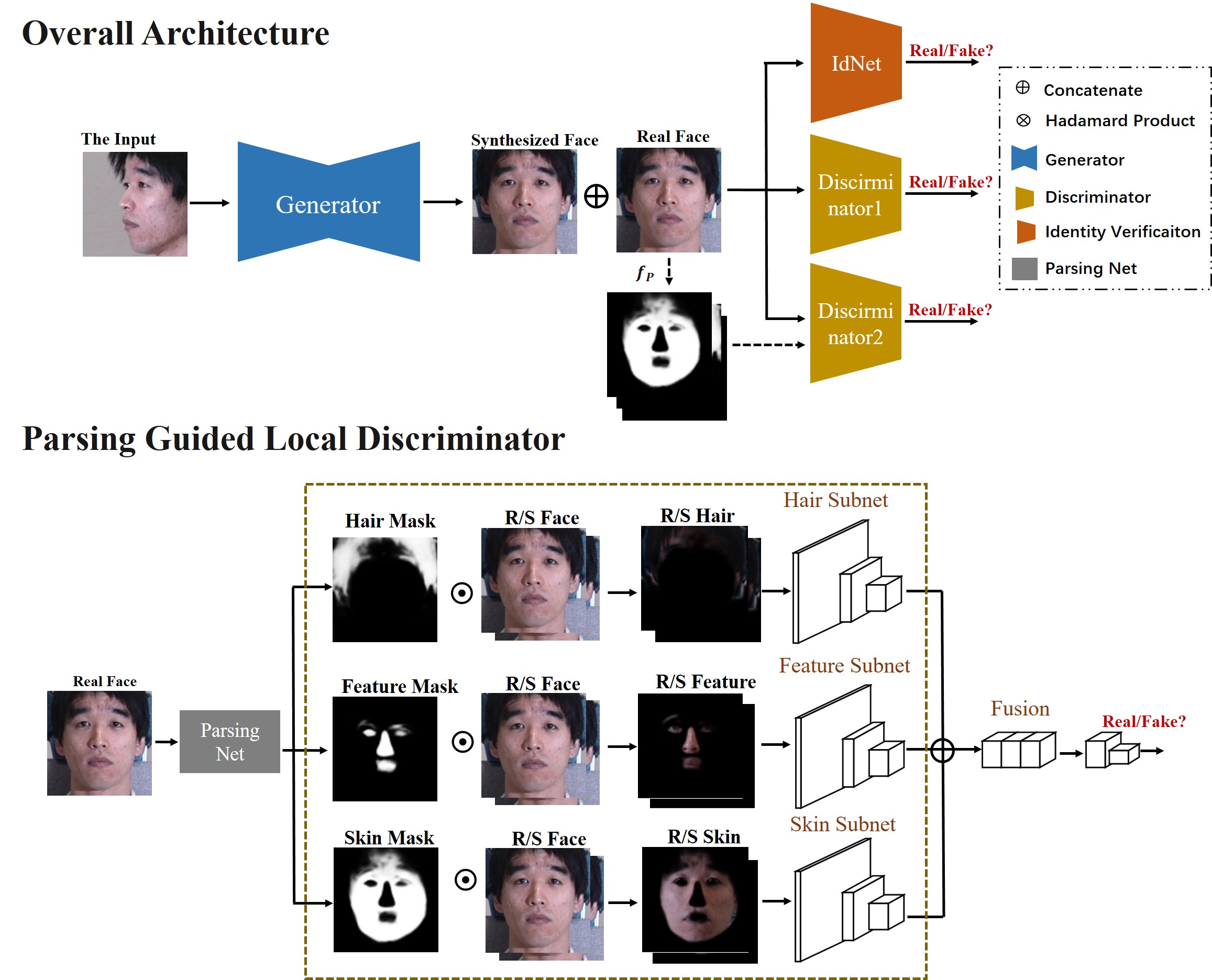}%images//GLA-GAN
\end{center}

\vspace{-0.4cm}
   \caption{The overall framework of our method.}\label{Fig:architecture_method}
\vspace{-0.4cm}
\end{figure}

As shown in Figure~\ref{Fig:architecture_method}, we introduce two discriminators during GAN optimization, including a global discriminator ${D_{{\theta _{D1}}}}$ and a parsing guided local discriminator ${D_{{\theta _{D2}}}}$. Specially, the discriminator ${D_{{\theta _{D1}}}}$ aims to distinguish the real
image $Y$ and the synthesized frontal image $\hat{Y}$ from a global view. Considering photo-realistic visualizations, especially
for faces with extreme yaw-pitch angles or accessory, it is
crucial to ensure the local consistency between the synthesized frontal image and the ground truth. First, we utilize a pre-trained facial parser $f_P$~\cite{liu2015multi} to capture three local masks, including the hairstyle mask ${M_h}$, the skin mask ${M_s}$ and the facial feature mask ${M_f}$ from the real frontal image $Y$,
\begin{equation}
\begin{array}{c}
{M_h},{M_s},{M_f} = {f_P}\left( Y \right)
\end{array}
\end{equation}
where the values of three masks are ranged from 0 to 1. Second, we treat these masks as the soft attention, facilitating the synthesized frontal image $\hat{Y}$ and the ground truth $Y$ as follows:
\begin{equation}
\begin{array}{c}
{Y_h} = Y \odot {M_h},{Y_s} = Y \odot {M_s},{Y_f} = Y \odot {M_f}
\end{array}
\end{equation}
\begin{equation}
\begin{array}{c}
{\hat{Y}_h} = \hat{Y} \odot {M_h},{\hat{Y}_s} = \hat{Y} \odot {M_s},{\hat{Y}_f} = \hat{Y} \odot {M_f}
\end{array}
\end{equation}
where $ \odot$ denotes the hadamard product. ${Y_h}$, ${Y_s}$ and ${Y_f}$ denote the hairstyle, skin and facial feature information from $Y$, while ${\hat{Y}_h}$, ${\hat{Y}_s}$ and ${\hat{Y}_f}$ are from $\hat{Y}$.
Then these local features are fed into the parsing guided local discriminator ${D_{{\theta _{D2}}}}$. As shown in Figure~\ref{Fig:architecture_method}, three subnets are used to encode the output feature maps of the hairstyle, skin and facial features, respectively. Finally, we concatenate the three encoded feature maps and feed it with binary cross entropy loss to distinguish that the input of local features is real or fake. The parsing guided local discriminator can efficiently ensure whether the local consistency of the synthesized frontal images is similar with the ground truth or
not.

\subsection{Training Losses}

\textbf{Multi-Scale Pixel Loss.} Following~\cite{hu2018pose}, we employ a multi-scale pixel loss to enhance the content consistency between the synthesized $\hat{Y}$ and the ground truth $Y$.
\begin{equation}
\begin{array}{c}
{L_{pixel}} = \frac{1}{3}\sum\limits_{i = 1}^3 {\frac{1}{{{W_i}{H_i}C}}\sum\limits_{w,h,c = 1}^{{W_i},{H_i},C} {\left| {\hat Y_{i,w,h,c} - Y_{i,w,h,c}} \right|} }
\end{array}
\end{equation}
where $C$ is the channel number, $i$ is the $i$-th image scale, $i \in \left\{ {1,2,3} \right\}$. $W_i$ and $H_i$ represent the width and height of the $i$-th image scale, respectively.

\textbf{Global-Local Adversarial Loss.} We adopt a global-local adversarial loss, aiming at synthesizing photo-realistic frontal face images. Specifically, the global discriminator ${D_{{\theta _{D1}}}}$ distinguishes the synthesized face image $\hat{Y}$ from real image $Y$.
\begin{equation}
\begin{array}{c}
\begin{array}{l}{L_{adv1}} = \mathop {\min }\limits_{{\theta _G}} \mathop {\max }\limits_{{\theta _{D1}}} {E_{Y\sim P(Y)}}[\log {D_{{\theta _{D1}}}}(Y)]\\\;\;\;\;\;\;\; + {E_{Y\sim P(Y)}}[\log (1 - {D_{{\theta _{D1}}}}(Y))]\end{array}
\end{array}
\end{equation}
The parsing guided local discriminator ${D_{{\theta _{D2}}}}$ aims to make the synthesized local facial details ${\hat{Y}_h}$, ${\hat{Y}_s}$ and ${\hat{Y}_f}$ close to the real ${Y_h}$, ${Y_s}$ and ${Y_f}$,
\begin{equation}\small
\resizebox{0.49\textwidth}{!}{
$\begin{array}{l}
\begin{array}{l}{L_{adv2}} = \mathop {\min }\limits_{{\theta _G}} \mathop {\max }\limits_{{\theta _{D2}}} {E_{_{{Y_h},{Y_s},{Y_f}\sim P({Y_h},{Y_s},{Y_f})}}}[\log {D_{{\theta _{D2}}}}({Y_h},{Y_s},{Y_f})]\\\;\;\;\;\; + {E_{{Y_h},{Y_s},{Y_f}\sim P({\hat{Y}_h},{\hat{Y}_s},{\hat{Y}_f})}}[\log (1 - {D_{{\theta _{D2}}}}({\hat{Y}_h},{\hat{Y}_s},{\hat{Y}_f}))]\end{array}
\end{array}$}
\end{equation}

\textbf{Identity Preserving Loss.} An identity preserving loss is employed to constrain the identity consistency between $\hat{Y}$ and $Y$. We utilize a pre-trained LightCNN-29~\cite{wu2018light} to extract the identity features from $\hat Y$ and $Y$. The identity preserving loss is as follows:
\begin{equation}
\begin{array}{l}
\begin{array}{l}{L_{id}} = {\rm{||}}{\varphi _f}{\rm{(Y)}} - {\varphi _f}{\rm{(\hat{Y})||}}_2^2\\\;\;\;\;\;\;\;{\rm{ + ||}}{\varphi _p}{\rm{(Y)}} - {\varphi _p}{\rm{(\hat{Y})||}}_F^2\end{array}
\end{array}
\end{equation}
where ${\varphi _f}$ and ${\varphi _p}$ denote the fully connected layer and the last pooling layer of the pre-trained LightCNN, respectively. \({\left\|  \cdot  \right\|_2}\) and \({\left\|  \cdot  \right\|_F}\) represent the vector 2-norm and matrix F-norm, respectively.

\textbf{Total Variation Regularization.} We introduce a total variation regularization term~\cite{johnson2016perceptual} to remove the unfavorable artifacts.
\begin{equation}
\label{e6}
{L_{tv}} = \sum\limits_{c = 1}^C {\sum\limits_{w,h = 1}^{W,H} {\left| {\hat Y_{w + 1,h,c} - \hat Y_{w,h,c}} \right| + \left| {\hat Y_{w,h + 1,c}^b - \hat Y_{w,h,c}} \right|} }
\end{equation}
where $C$, $W$ and $H$ are the channel, width and height of the synthesized image $Y$, respectively.

\textbf{Overall Loss.} Finally, the total supervised loss is a weighted sum of the above losses. The generator and two discriminators, including a global discriminator and a parsing guided local discriminator, are trained alternately to play a min-max problem. The overall loss is written as:
\begin{equation}
L = {\lambda _1}{L_{pixel}} + {\lambda _2}{L_{adv1}} + {\lambda _3}{L_{adv2}} + {\lambda _4}{L_{id}} + {\lambda _5}{L_{tv}}
\end{equation}
where $\lambda _1, \lambda _2, \lambda _3$, $\lambda _4$ and $\lambda _5$ are the trade-off parameters.

%-----------------------------------------------------------
\section{Experiments}
We evaluate our method qualitatively and quantitatively
on the proposed M$^{2}$FPA database. For qualitative evaluation, we show the results of face frontalization on several yaw and pitch faces. For quantitative evaluation, we perform pose-invariant face recognition based on both the original and synthesized face images. We also provide three face frontalization benchmarks on  M$^{2}$FPA, including DR-GAN \cite{tran2017disentangled}, TP-GAN \cite{huang2017beyond} and CAPG-GAN \cite{hu2018pose}. To further demonstrate the effectiveness of the proposed method and assess the difficulty of  M$^{2}$FPA, we also conduct experiments on Multi-PIE \cite{gross2010multi} database, which is widely used in facial pose analysis. In the following subsections, we begin with an introduction of databases and settings, especially the training and testing protocols of  M$^{2}$FPA. Then we present the qualitative frontalization results and quantitative recognition results on  M$^{2}$FPA and Multi-PIE. Lastly, we conduct ablation study to demonstrate the effect of each part in our method.

\subsection{Databases and Settings}
\textbf{Databases.} The M$^{2}$FPA database totally contains 397,544 images of 229 subjects under 62 poses, 4 attributes and 7 illuminations. 57 of 62 poses are chosen in our experiments, except for $+45^\circ$ pitch angles. We randomly select 162 subjects as the training set, i.e., $162 \times 57 \times 4 \times 7 = 258,552$ images in total. The remaining 67 subjects form the testing set. For testing, one gallery image with frontal view, neutral attribute and above illumination is employed for each of the 67 subjects. The remaining yaw and pitch face images are treated as probes. The number of the probe and gallery images are 105,056 and 67 respectively. We will release the original M$^{2}$FPA database together with the annotated five facial landmarks and the training and testing protocols.

The Multi-PIE database \cite{gross2010multi} is a popular database for evaluating face synthesis and recognition across yaw angles. Following \cite{hu2018pose}, we use Setting 2 protocol in our experiments. There are 161,460, 72,000, 137 images in the training, probe and gallery sets, respectively.

\textbf{Implementation Details.} Following the previous methods \cite{tran2017disentangled,huang2017beyond,hu2018pose}, we crop and align $128 \times 128$ face images on M$^{2}$FPA and Multi-PIE for experimental evaluation. Besides, we also conduct experiments on $256 \times 256$ face images on M$^{2}$FPA for high-resolution face frontalization under multiple yaw and pitch variations. A pre-trained LightCNN-29 \cite{wu2018light} is chosen for calculating the identity preserving loss and is fixed during training. Our model is implemented with Pytorch.  We choose Adam optimizer with the ${\beta _1}$ of 0.5 and ${\beta _2}$ of 0.99. The learning rate is initialized by $2{e{ - 4}}$ and linearly decayed by $2{e{ - 5}}$ after each epoch until 0. The batch size is 16 for $128 \times 128$ resolution and 8 for $256 \times 256$ resolution on a single NVIDIA TITAN Xp GPU with 12G memory. In all experiments, we empirically set the trade-off parameters $\lambda _1, \lambda _2, \lambda _3$, $\lambda _4$ and $\lambda _5$ to 20, 1, 1, 0.08 and $1{e{ - 4}}$, respectively.

\subsection{Evaluation on M$^{2}$FPA}
\subsubsection{Face Frontalization}

\begin{figure*}[t]
\setlength{\abovecaptionskip}{0cm}
\begin{center}
\includegraphics[width=0.85\linewidth]{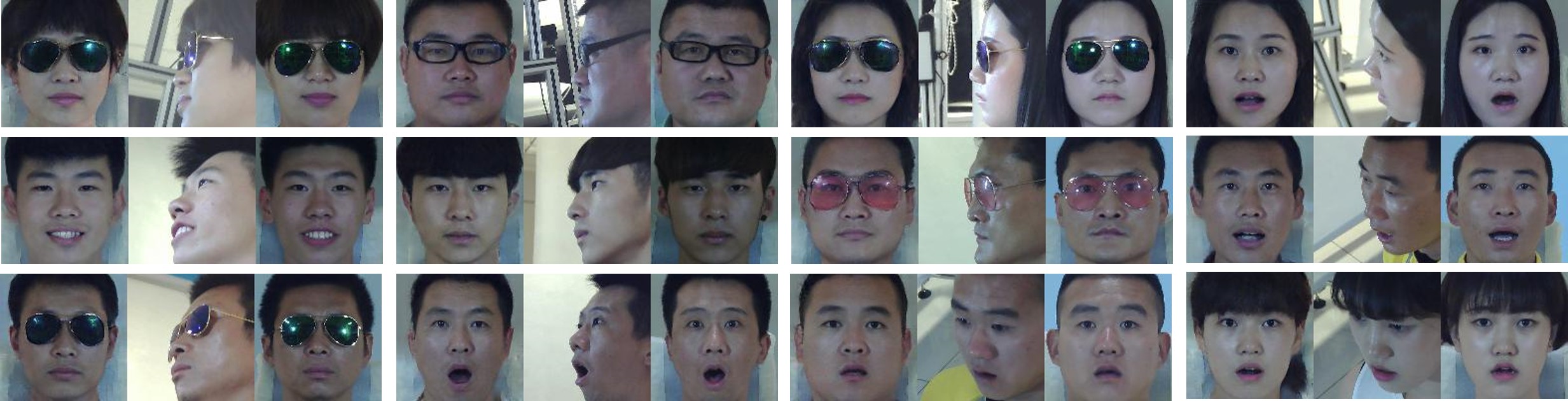}%images//GLA-GAN
\end{center}

\vspace{-0.4cm}
   \caption{The frontalized 128$\times$128 results of our method under different poses on M$^{2}$FPA. From top to bottom, the yaw angles are $+90^\circ$, $+75^\circ$, and $+60^\circ$. For each subject, the first column is the generated frontal image, the second column is the input profile, and the last column is the ground-truth frontal image.}\label{Fig:result_of_ourmethod}
\vspace{-0.4cm}
%\label{fig:synthesis_results}
\end{figure*}
\begin{figure*}[t]
\setlength{\abovecaptionskip}{0cm}
\begin{center}
\includegraphics[width=0.85\linewidth]{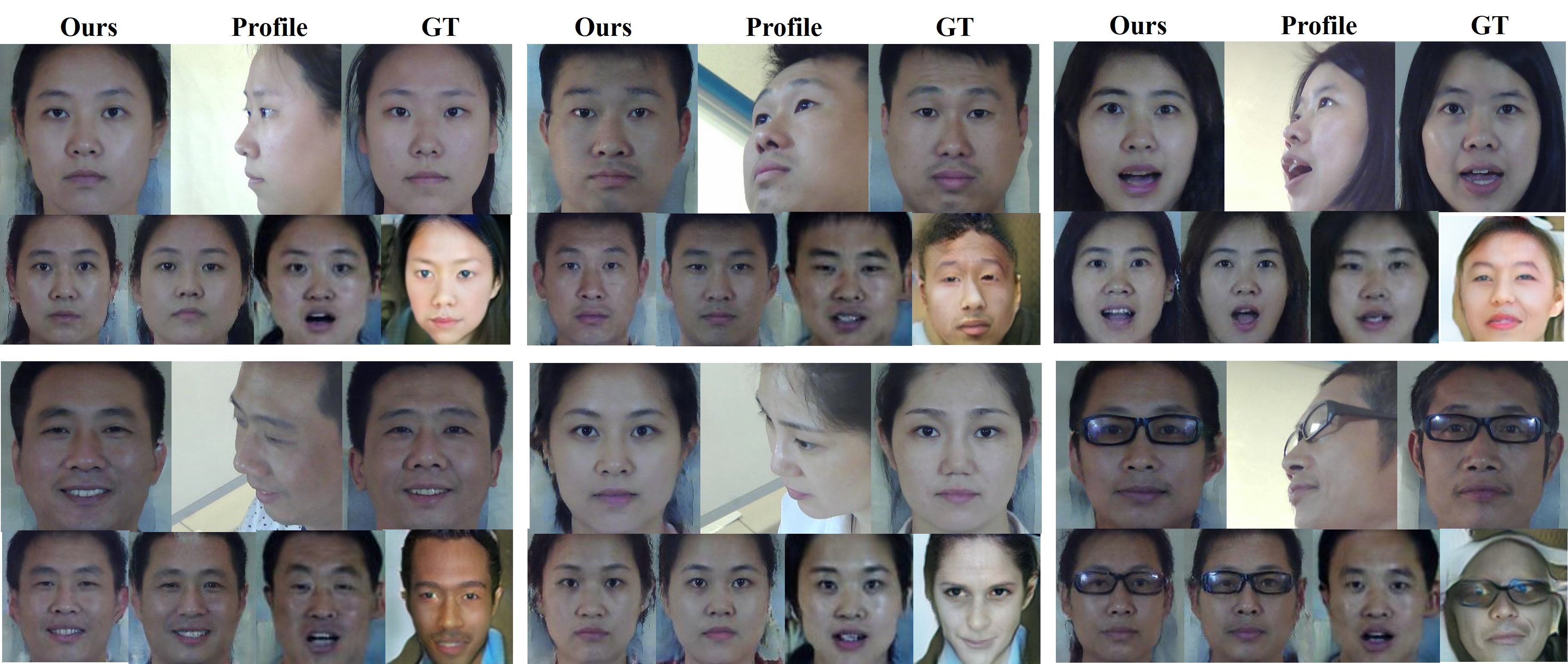}%images//GLA-GAN
\end{center}

\vspace{-0.4cm}
   \caption{Frontalized results of different methods under extreme poses on M$^{2}$FPA. For each subject, the first row shows the visualizations (256$\times$256) of our method. From left to right: our frontalized result, the input profile and the groundtruth. The second row shows the frontalized results (128$\times$128) of different benchmark methods. From left to right: CAPG-GAN \cite{hu2018pose}, TP-GAN \cite{huang2017beyond}, DR-GAN \cite{tran2017disentangled} (96$\times$96) and the online demo.
   }\label{Fig:result_of_ourmethod256}
\vspace{-0.4cm}
%\label{fig:synthesis_results}
\end{figure*}

The collected M$^{2}$FPA database provides a possibility for face frontalization under various yaw and pitch angles. Benefiting from the global-local adversary, our method can frontalize facial images with large yaw and pitch angles. The synthesis results of $+60^\circ$$\sim$$+90^\circ$ yaw angles and $-30^\circ$$\sim$$+30^\circ$ pitch angles are shown in Figure \ref{Fig:result_of_ourmethod}. We observe that not only the global facial structure but also the local texture details are recovered in an identity consistent way. Surprisingly, the sunglasses under extreme poses can also be well preserved. Besides, the current databases for large pose face frontalization are limited to yaw angles and a low resolution, i.e. $128 \times 128$. The collected M$^{2}$FPA has higher quality and supports for face frontalization at $256 \times 256$ resolution with multiple yaw and pitch angles. The frontalized $256 \times 256$ results of our method on M$^{2}$FPA are presented in Figure \ref{Fig:result_of_ourmethod256}, where high quality and photo-realistic frontal faces are obtained. More frontalized results are listed in supplementary materials.

In addition, we provide several benchmark face frontalization results on M$^{2}$FPA, including DR-GAN\cite{tran2017disentangled}, TP-GAN\cite{huang2017beyond}, and CAPG-GAN\cite{hu2018pose}. We re-implement CAPG-GAN and TP-GAN according to the original papers. For DR-GAN, we provide two results: one is the re-implemented version$\footnote{\url{https://github.com/zhangjunh/DR-GAN-by-pytorch}}$ and the other is the online demo$\footnote{\url{http://cvlab.cse.msu.edu/cvl-demo/DR-GAN-DEMO/index.html}}$. Figure \ref{Fig:result_of_ourmethod256} presents the comparison results. We observe that our method, CAPG-GAN and TP-GAN achieve good visualizations, while DR-GAN fails to preserve the attributes and the facial structures due to its unsupervised learning procedure. However, there are also some unsatisfactory synthesized details among most of the methods, such as the hair, the face shape. These demonstrate the difficulties of synthesizing photorealistic frontal faces from extreme yaw and pitch angles. Therefore, we expect that collected M$^{2}$FPA pushes forward the advance in multiple yaw and pitch face synthesis.

\subsubsection{Pose-invariant Face Recognition}
\begin{table}\scriptsize
\centering
\caption{Rank-1 recognition rates ($\%$) across views at $0^\circ$ pitch angle on M$^{2}$FPA.}
\begin{tabular}{lllllllllll} \toprule[1pt]
 Method&$\pm15^\circ$&$\pm30^\circ$&$\pm45^\circ$&$\pm60^\circ$&$\pm75^\circ$&$\pm90^\circ$\\
%\midrule[0.7pt]
\cline{1-7}
\cline{1-7}
\multicolumn{7}{c}{ \textbf{LightCNN-29 v2}}\\
\cline{1-7}
Original & 100 & 100 & 99.8 & 98.6 & 86.9 & 51.7\\

DR-GAN\cite{tran2017disentangled} &98.9&97.9&95.7&89.5&70.3&35.5\\

TP-GAN\cite{huang2017beyond} &99.9&99.8&99.4&97.3&87.6&62.1\\

CAPG-GAN\cite{hu2018pose} &99.9&99.7&99.4&96.4&87.2&63.9\\

Ours & \textbf{100}&\textbf{100}&\textbf{99.9}&\textbf{98.4}&\textbf{90.6}&67.6\\
%\midrule[0.7pt]
\cline{1-7}
\cline{1-7}
\multicolumn{7}{c}{ \textbf{IR-50}}\\
\cline{1-7}
Original & 99.7 & 99.7 & 99.2 & 97.2 & 87.2 & 35.3\\

DR-GAN\cite{tran2017disentangled} &97.8&97.6&95.6&89.9&70.6&26.5\\

TP-GAN\cite{huang2017beyond} &99.7&99.2&98.2&96.3&86.6&48.0\\

CAPG-GAN\cite{hu2018pose} &98.8&98.5&97.0&93.4&81.9&50.1\\

Ours & \textbf{99.5}&\textbf{99.5}&\textbf{99.0}&\textbf{97.3}&\textbf{89.6}&\textbf{55.8}\\
\bottomrule[1pt]
 %TP-GAN+align & 62.43 & 76.69 & 87.63 & - & - & - \\
   \end{tabular}\label{tab:pitch0}
   \vspace{-0.4cm}
\end{table}

\begin{table}\scriptsize
\centering
\caption{Rank-1 recognition rates ($\%$) across views at $\pm15^\circ$ pitch angle on M$^{2}$FPA.}
\begin{tabular}{p{1.2cm}|p{0.43cm}|p{0.35cm}p{0.4cm}p{0.4cm}p{0.4cm}p{0.4cm}p{0.4cm}p{0.4cm}}
\toprule[1pt]

Method&Pitch&$\pm0^\circ$&$\pm15^\circ$&$\pm30^\circ$&$\pm45^\circ$&$\pm60^\circ$&$\pm75^\circ$&$\pm90^\circ$\\
\cline{1-9}
\cline{1-9}
\multicolumn{9}{c}{ \textbf{LightCNN-29 v2}}\\
\cline{1-9}
 \multirow{2}{*}{Original} &$+15^\circ$& 100 & 100 & 100 & 99.8 & 97.5 & 76.5&34.3\\
 %\cline{2-9}
&$-15^\circ$& 99.9& 100 & 99.8 &99.7 & 97.3 &81.8&45.9\\
\cline{1-9}
\multirow{2}{*}{DR-GAN\cite{tran2017disentangled}}&$+15^\circ$&99.1&98.8&98.0&94.8&85.6&61.1&20.8\\
 %\cline{2-9}
&$-15^\circ$&98.1&98.2&96.5&93.3&83.1&62.7&31.0\\
\cline{1-9}
\multirow{2}{*}{TP-GAN\cite{huang2017beyond}}&$+15^\circ$&99.8&99.8&99.7&99.5&95.7&81.6&50.9\\
 %\cline{2-9}
&$-15^\circ$&99.9&99.9&99.6&99.2&95.9&84.1&56.9\\
\cline{1-9}
\multirow{2}{*}{CAPG-GAN}&$+15^\circ$&99.8&99.9&99.8&98.9&95.0&81.4&54.4\\
 %\cline{2-9}
\multicolumn{1}{c|}{\cite{hu2018pose}}&$-15^\circ$&99.8&99.9&99.7&98.7&95.1&85.5&65.6\\
\cline{1-9}
\multirow{2}{*}{Ours }&$+15^\circ$& \textbf{99.9}& \textbf{99.9}&\textbf{99.8}&\textbf{99.7}&\textbf{97.5}&\textbf{86.2}&\textbf{56.2}\\
&$-15^\circ$& \textbf{99.9}&
\textbf{99.9}&\textbf{99.8}&\textbf{99.7}&\textbf{97.4}&\textbf{88.1}&\textbf{66.5}\\
\midrule[0.7pt]
\multicolumn{9}{c}{ \textbf{IR-50}}\\
\cline{1-9}
 \multirow{2}{*}{Original} &$+15^\circ$& 99.8 & 99.9 & 99.6 & 98.7 & 95.7& 77.1&23.4\\
&$-15^\circ$& 98.7& 99.4 & 99.2 &98.1 & 95.7 &78.8&27.9\\
\cline{1-9}

\multirow{2}{*}{DR-GAN\cite{tran2017disentangled}}&$+15^\circ$&98.5&98.2&97.8&94.0&84.8&60.9&17.0\\
&$-15^\circ$&95.8&97.2&96.2&93.3&84.8&60.3&20.8\\
\cline{1-9}

\multirow{2}{*}{TP-GAN\cite{huang2017beyond}}&$+15^\circ$&99.0&99.6&99.1&98.5&94.7&79.1&40.6\\
&$-15^\circ$&98.2&98.9&98.1&97.2&94.8&80.9&43.5\\
\cline{1-9}

\multirow{2}{*}{CAPG-GAN}&$+15^\circ$&98.9& 99.0&98.5&95.8&91.5&75.7&40.7\\
\multicolumn{1}{c|}{\cite{hu2018pose}}&$-15^\circ$&98.5&98.5&97.9&95.3&90.3&76.0&47.8\\
\cline{1-9}

\multirow{2}{*}{Ours }&$+15^\circ$& \textbf{99.7}& \textbf{99.6}&\textbf{99.4}&\textbf{98.7}&\textbf{96.1}&\textbf{84.5}&\textbf{43.6}\\
&$-15^\circ$& \textbf{98.6}& \textbf{99.1}&\textbf{98.7}&\textbf{98.8}&\textbf{96.5}&\textbf{83.9}&\textbf{49.7}\\
\bottomrule[1pt]

 %TP-GAN+align & 62.43 & 76.69 & 87.63 & - & - & - \\
   \end{tabular}\label{tab:pitch15}
   \vspace{-0.4cm}
\end{table}

\begin{table}\scriptsize
\centering
\caption{Rank-1 recognition rates ($\%$) across views at $\pm30^\circ$ pitch angle on M$^{2}$FPA.}
\begin{tabular}{l|c|cccccccccccccc}\toprule[1pt]
 Method&Pitch&$\pm0^\circ$&$\pm22.5^\circ$&$\pm45^\circ$&$\pm67.5^\circ$&$\pm90^\circ$\\
\cline{1-7}
\cline{1-7}
\multicolumn{7}{c}{ \textbf{LightCNN-29 v2}}\\
\cline{1-7}
\multirow{2}{*}{Original} &$+30^\circ$& 99.7& 99.2& 96.5 & 71.6&24.5 \\
&$-30^\circ$&98.6&98.2 & 93.6&69.9&22.1\\
\cline{1-7}
\multirow{2}{*}{DR-GAN\cite{tran2017disentangled}}&$+30^\circ$&93.8&91.5&83.4&52.0&16.9\\
&$-30^\circ$&91.7&90.6&79.1&46.6&16.6\\
\cline{1-7}
\multirow{2}{*}{TP-GAN\cite{huang2017beyond}}&$+30^\circ$&99.7&98.8&95.8&77.2&43.4\\
&$-30^\circ$&98.2&97.6&93.4&75.7&38.9\\
\cline{1-7}
\multirow{2}{*}{CAPG-GAN\cite{hu2018pose}}&$+30^\circ$&98.8&98.4&94.1&79.5&48.0\\
&$-30^\circ$&98.9&98.3&93.8&75.3&49.3\\
\cline{1-7}
\multirow{2}{*}{Ours }&$+30^\circ$& \textbf{99.7}& \textbf{99.1}&\textbf{97.7}&\textbf{81.9}&\textbf{48.2}\\
&$-30^\circ$& \textbf{98.9}& \textbf{98.7}&\textbf{95.8}&\textbf{82.2}&\textbf{49.3}\\
\midrule[0.7pt]
\multicolumn{7}{c}{ \textbf{IR-50}}\\
\cline{1-7}
\multirow{2}{*}{Original} &$+30^\circ$& 99.2& 98.1& 94.7 & 73.5&17.6\\
&$-30^\circ$&97.1&97.3 & 93.0&67.2&9.0\\
\cline{1-7}
\multirow{2}{*}{DR-GAN\cite{tran2017disentangled}}&$+30^\circ$&92.9&92.3&83.8&56.4&13.9\\
&$-30^\circ$&93.0&92.0&82.1&50.3&7.5\\
\cline{1-7}
\multirow{2}{*}{TP-GAN\cite{huang2017beyond}}&$+30^\circ$&98.1&97.3&94.4&76.8&34.5\\
&$-30^\circ$&95.7&96.1&92.2&71.6&27.5\\
\cline{1-7}
\multirow{2}{*}{CAPG-GAN\cite{hu2018pose}}&$+30^\circ$&97.1&96.2&90.5&73.1&34.5\\
&$-30^\circ$&95.8&95.4&89.2&67.6&33.0\\
\cline{1-7}
\multirow{2}{*}{Ours }&$+30^\circ$& \textbf{98.6}& \textbf{97.8}&\textbf{96.0}&\textbf{79.6}&\textbf{36.4}\\
&$-30^\circ$& \textbf{97.2}& \textbf{97.4}&\textbf{95.1}&\textbf{76.7}&\textbf{33.1}\\
\bottomrule[1pt]
 %TP-GAN+align & 62.43 & 76.69 & 87.63 & - & - & - \\
   \end{tabular}\label{tab:pitch30}
   \vspace{-0.4cm}
\end{table}

Face recognition accuracy is a commonly used metric to evaluate the identity preserving ability of different frontalization methods. The better the recognition accuracy, the more identity information is preserved during the synthesis process. Hence, we quantitatively evaluate our method and compare it with several state-of-the-art frontalization methods on M$^{2}$FPA, including DR-GAN\cite{tran2017disentangled}, TP-GAN\cite{huang2017beyond}, and CAPG-GAN\cite{hu2018pose}. We employ two open-source pre-trained recognition models, LightCNN-29 v2$\footnote{\url{https://github.com/AlfredXiangWu/LightCNN}}$ and IR-50$\footnote{\url{https://github.com/ZhaoJ9014/face.evoLVe.PyTorch}}$, as the feature extractors and define the distance metric as the average distance between the original image pair and the generated image pair. Tables \ref{tab:pitch0}, \ref{tab:pitch15} and \ref{tab:pitch30} present the Rank-1 accuracies of different methods on M$^{2}$FPA under $0^\circ$, $\pm15^\circ$ and $\pm30^\circ$ pitch angles, respectively. When keeping the yaw angle consistent, we observe that the larger the pitch angle, the lower the accuracy is obtained, suggesting the great challenge in pitch variations.  Besides, by ¡°recognition via generation¡±, TP-GAN, CAPG-GAN and our method achieve better recognition performance than the original data under the large poses, such as $\pm90^\circ$ yaw and $\pm30^\circ$ pitch angles. We further observe that the accuracy of DR-GAN is inferior to the original data. The reason may be that DR-GAN is trained in an unsupervised way and there are too many pose variations in M$^{2}$FPA.

\subsection{Evaluation on Multi-PIE}

\begin{figure}[t]
\setlength{\abovecaptionskip}{0cm}
\begin{center}
\includegraphics[width=0.9\linewidth]{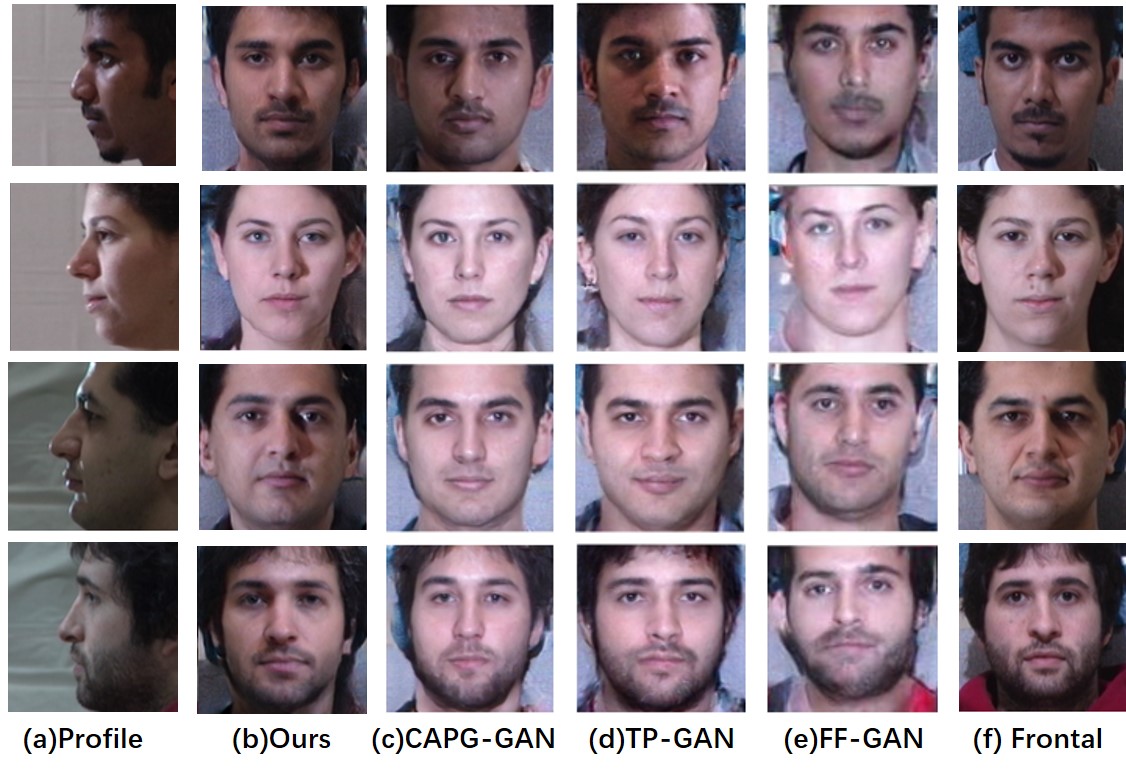}%images//GLA-GAN
\end{center}

\vspace{-0.4cm}
   \caption{Comparisons with different methods under the pose of $75^\circ$(first two rows) and $90^\circ$(last two rows) on Multi-PIE.}\label{Fig:compare_multipie}
\vspace{-0.3cm}
%\label{fig:synthesis_results}
\end{figure}

In this section, we present the quantitative and qualitative evaluations on the popular Multi-PIE \cite{gross2010multi} database. Figure~\ref{Fig:compare_multipie} shows the frontalized image of our method. We observe that our method can achieve photo-realistic visualizations against other state-of-the-art methods, including CAPG-GAN~\cite{hu2018pose}, TP-GAN~\cite{huang2017beyond} and FF-GAN~\cite{yin2017towards}. Table \ref{tab:set2} further tabulates the Rank-1 performance of different methods under the Setting 2 for Multi-PIE. It is obvious that our method outperforms its competitors, including  FIP+LDA\cite{zhu2013deep}, MVP+LDA\cite{zhu2014multi}, CPF\cite{yim2015rotating}, DR-GAN\cite{tran2017disentangled}, FF-GAN\cite{yin2017towards}, TP-GAN\cite{huang2017beyond} and CAPG-GAN\cite{hu2018pose}.

\begin{table}\scriptsize
\centering
\caption{Rank-1 recognition rates ($\%$) across views under Setting 2 on Multi-PIE.}
\begin{tabular}{lclclclclcl} \toprule[1pt]
 Method&$\pm15^\circ$&$\pm30^\circ$&$\pm45^\circ$&$\pm60^\circ$&$\pm75^\circ$&$\pm90^\circ$\\
 \hline \hline
 FIP+LDA\cite{zhu2013deep} & 90.7 & 80.7 & 64.1 & 45.9 & - & -\\
MVP+LDA\cite{zhu2014multi} & 92.8 & 83.7  & 72.9 & 60.1 & - & - \\
CPF\cite{yim2015rotating} & 95.0 & 88.5 & 79.9 & 61.9 & - & -\\
DR-GAN\cite{tran2017disentangled}  & 94.0 & 90.1 & 86.2 & 83.2 & - & -\\
FF-GAN\cite{yin2017towards} & 94.6 & 92.5 &89.7 & 85.2 & 77.2 & 61.2\\
TP-GAN\cite{huang2017beyond} & 98.68 & 98.06 & 95.38 & 87.72 & 77.43 & 64.64\\
CAPG-GAN\cite{hu2018pose} & 99.82&99.56 & 97.33 & 90.63 & 83.05 & 66.05\\
%3D-PIM\cite{zhao20183d} & 99.64& 99.48 &98.81 & 98.37 & 95.21& 86.73\\
\hline
Ours & 99.96&99.78&99.53&96.18&88.74&75.33\\
\bottomrule[1pt]
   \end{tabular}\label{tab:set2}
\vspace{-0.4cm}
\end{table}

%-------------------------------------------------------------------------
\subsection{Ablation Study}

We report both quantitative recognition results and qualitative visualization results of our method and its four variants for a comprehensive comparison as the ablation study. We give the details in the Supplemental Materials.

\section{Conclusion}
This paper has introduced a new large-scale \textbf{M}ulti-yaw \textbf{M}ulti-pitch high-quality database for \textbf{F}acial \textbf{P}ose \textbf{A}nalysis (M$^2$FPA), including face frontalization, face rotation, facial pose estimation and pose-invariant face recognition. To the best of our knowledge, M$^2$FPA is the most comprehensive multi-view face database that covers variations in yaw, pitch, attribute, illumination, accessory. We also provide an effective benchmark for face frontalization and pose-invariant face recognition on M$^2$FPA. Several state-of-the-art methods, such as DR-GAN, TP-GAN and CAPG-GAN, are implemented and evaluated. Moreover, we propose a simple yet effective parsing guided local discriminator to capture the local consistency during GAN optimization. In this way, we can synthesize photo-realistic frontal images with extreme yaw and pitch variations on Multi-PIE and M$^2$ FPA. We believe that the new database and benchmark can significantly push forward the advance of facial pose analysis in community.
\section{Acknowledgement}
This work is partially funded by National Natural Science Foundation of China (Grant No. 61622310, U1836217, 61427811) and Beijing Natural Science Foundation (Grant No. JQ18017).
{\small
\bibliographystyle{ieee_fullname}
\bibliography{egbib}
}

\end{document}